\documentclass[letterpaper]{article} 
\usepackage{aaai2027}  
\usepackage[hyphens]{url}  
\usepackage{graphicx} 
\usepackage{natbib}  
\usepackage{caption} 
\usepackage{booktabs}
\usepackage{amsmath}
\usepackage{amssymb}
\usepackage{algorithm}
\usepackage{algorithmic}
\usepackage{array}
\newcolumntype{L}[1]{>{\raggedright\arraybackslash}p{#1}}

\ifdefined\pdfinfoomitdate
\fi
\ifdefined\pdftrailerid
  \pdftrailerid{}
\fi
\ifdefined\pdfsuppressptexinfo
\fi
\title{Think Only When Needed: Prompt-Authority Control for Selective Slow-Path Intervention in Vision-Language-Action Manipulation}
\author{
Zhiruo Zhou\textsuperscript{1,2,*},
Zelin Li\textsuperscript{1,*,\textdagger},
Xiwen Chen\textsuperscript{1,3},
Jiazhuo Li\textsuperscript{4},
Chenwei Wang\textsuperscript{5},
Huiming Chen\textsuperscript{6}
Xiaojun Zhu\textsuperscript{1,\textdagger}
}
\affiliations{
\textsuperscript{1}SIGS, Tsinghua University, China; 
\textsuperscript{2}Wuhan University of Technology, China;
\textsuperscript{3}Shanghai Jiao Tong University, China; 
\textsuperscript{4}University of Michigan - Ann Arbor, USA;
\textsuperscript{5}AiDlab, Hong Kong (China SAR);
\textsuperscript{6}City University of Hong Kong, Hong Kong (China SAR);
\\
\textsuperscript{*}Equal Contributions; \textsuperscript{\textdagger}Correspondence: chiellini.lee@gmail.com; 
zhu.xiaojun@sz.tsinghua.edu.cn
}

\newcommand{\ck}{\checkmark}
\newcommand{\na}{--}
\newcommand{\pipizero}{$\pi_0$}
\newcommand{\pizerofive}{$\pi_{0.5}$}

\begin{document}

\maketitle

\begin{abstract}
Retrieval can efficiently and effectively augment a frozen vision--language--action (VLA) policy without
retraining, yet retrieved text becomes a control intervention once it enters
the executed prompt. In a matched audit, raw appended text reduces mean success
from 92.47\% to 3.00\%, while meaningful and length-matched meaningless
appends both fail on all 500 states. This result identifies
\emph{prompt-form collapse}: changing the instruction form, rather than adding
useful semantics, can dominate execution. We introduce TOWN-VLA (Think Only
When Needed), a prompt-authority interface that separates candidate generation from
permission to alter the policy input. A fixed compatibility rule authorizes a
canonical compact instruction; otherwise, the interface restores the original
Base prompt exactly. Across 900 audited routes, every route follows this
contract: 525 routes recover Base with matching hashes, and all 375 authorized
prompts preserve the task signature. On a matched $4\times7$ LIBERO-Plus
evaluation with 10{,}030 episodes per method, success rises from 69.5\% to
73.1\% ($+362$ episodes; 95\% CI 1.89--5.45 points), improving on six
perturbation axes and all four suites. On a physical PiPER arm with a frozen
\pizerofive{} checkpoint, success rises from 52.7\% to 78.7\% over 150 trials
per method ($p=3.16\times10^{-6}$). Prompt authority is enforceable for a frozen controller; oracle-free admission calibration is the next deployment target.
\end{abstract}

\section{Introduction}

Vision-language-action (VLA) policies map visual observations and language
instructions to closed-loop robot actions, connecting the semantic breadth of
vision-language pretraining with continuous control
\cite{driess2023palme,brohan2023rt2,kim2024openvla}. Cross-embodiment data,
generalist pretraining, and continuous-action models have broadened their
capabilities \cite{openx2023,octo2024,black2024pi0,liu2024rdt,pertsch2025fast}.
Practical systems now wrap frozen policies with planners, memories, verifiers,
or steering modules that supply test-time guidance
\cite{ahn2022saycan,huang2022inner,zawalski2025ecot,li2025mapvla}. Once such
text enters the executed instruction, it becomes a control intervention. We study this interface rather than policy retraining, asking which retrieved
content may modify the frozen policy input (Figures 1 and 2).

\begin{figure}[!t]
\centering
\includegraphics[width=0.95\columnwidth]{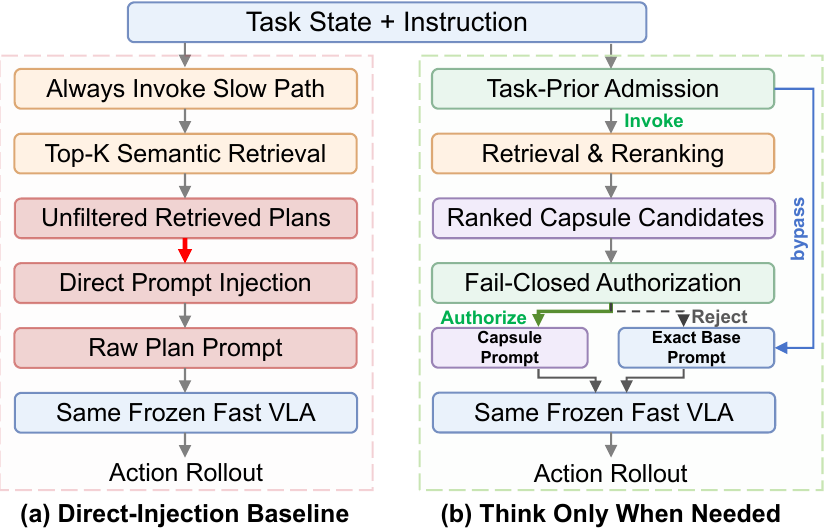}
\caption{Prompt authority at the frozen-policy boundary. Direct injection
passes retrieved text to the VLA; TOWN-VLA (Think Only When Needed) authorizes a canonical capsule
or restores Base while retaining the same frozen action generator.}
\label{fig:teaser}
\end{figure}

Holding the policy, initial states, and execution protocol fixed, raw appended
text reduces mean success from 92.47\% to 3.00\%
(Figure~\ref{fig:harm}(a)). In a paired 500-state control, meaningful and
length-matched meaningless appends both yield 0/500, called
\emph{prompt-form collapse}. We hypothesize that OFT (Optimized Fine-Tuning) adaptation exposes the
policy to a stable instruction template but no appended retrieval tokens;
appending text shifts sequence length, token positions, and the instruction
boundary. The identical failure of semantically different appends, while
canonical prompts remain operative, supports template departure over semantic
quality, demonstrating task-preserving rewordings and
multimodal perturbations \cite{srikanth2026qdig,xie2026strongvla} by isolating
the interface around an otherwise frozen policy.

Existing planners ground proposals through affordances, feedback, programs,
or spatial value maps \cite{ahn2022saycan,huang2022inner,liang2022code,huang2023voxposer};
newer systems request assistance, route experts, retrieve memory, or steer
frozen policies
\cite{ren2023askhelp,zhang2026corevla,ren2026routervla,li2025mapvla,zhang2026harnessvla,jeong2026mostlyharmless}.
They make the proposal-to-action interface consequential, but do not generally
separate candidate quality from an explicit, logged authorization decision.
This distinction parallels value-of-computation reasoning and selective
prediction with a reject option
\cite{russell1991metareasoning,geifman2019selectivenet}: a slow path may be
worth querying without its output being worth obeying.We therefore formulate retrieval augmentation as
\emph{prompt-authority control} in TOWN-VLA. Figure~\ref{fig:teaser}
contrasts direct injection with rejection followed by Base-prompt restoration.
Memory is one replaceable source of proposals; the validated method concerns
the contract governing whether and how any candidate may alter the frozen
policy input, rather than attributing gains to retrieved strategy content.

TOWN-VLA separates candidate preparation from prompt authorization
(Figure 2). A Compatibility-Reranked Capsule retrieves structured candidates
and orders them by a fixed text-level compatibility rule. A Top-2 Fail-Closed
Cascade inspects at most two candidates: the first eligible candidate becomes
a canonical compact instruction; if neither passes, the interface retains the
exact Base Policy prompt. The frozen Base Policy remains the sole action
generator, with its parameters and control loop unchanged. The construction
supplies two verifiable properties: (P1) every unauthorized route resolves to
the bit-identical Base prompt, and (P2) each inspected episode uses at most one
retrieval, five compatibility scores, and two checks. Task-Prior Admission
estimates an oracle-side upper bound on removable slow-path computation;
oracle-free admission is evaluated in Q4.

\paragraph{Key Contributions:}
\begin{itemize}

  \item We formulate \emph{prompt-authority control} as a previously
underexplored test-time interface problem for frozen VLAs and identify \emph{prompt-form collapse} as its motivating failure mode. Raw appended text reduces mean
success from 92.47\% to 3.00\%; task-aligned and length-matched
meaningless suffixes both yield 0/500, whereas exact Base restoration
reaches 499/500.
    \item We operationalize \emph{proposal $\neq$ authority} at the
    frozen-policy boundary: TOWN-VLA admits only canonical guidance,
    restores rejected routes to the exact Base prompt, and limits inspection
    to Top-2. Across 900 routes, 525 recover hash-identical Base prompts and
    all 375 authorized prompts preserve the task signature.
    
  \item In matched evaluation, TOWN-VLA raises LIBERO-Plus success from
69.46\% to 73.07\%
  ($+362$ successes; paired-cell bootstrap 95\% CI, 1.89--5.45 points), with
  gains on six of seven axes and all four suites; physical PiPER success rises
  from 52.7\% to 78.7\%.

  \item A preregistered boundary quantifies admission timing. Oracle
routing preserves 2{,}826 successes, halving slow-path calls; on the fixed
24/36 development--held-out split, oracle-free gates peak at the
91.81\% Base Policy. Admission calibration is the next challenge.
\end{itemize}

\begin{figure*}[t]
\centering
\includegraphics[width=0.95\textwidth]{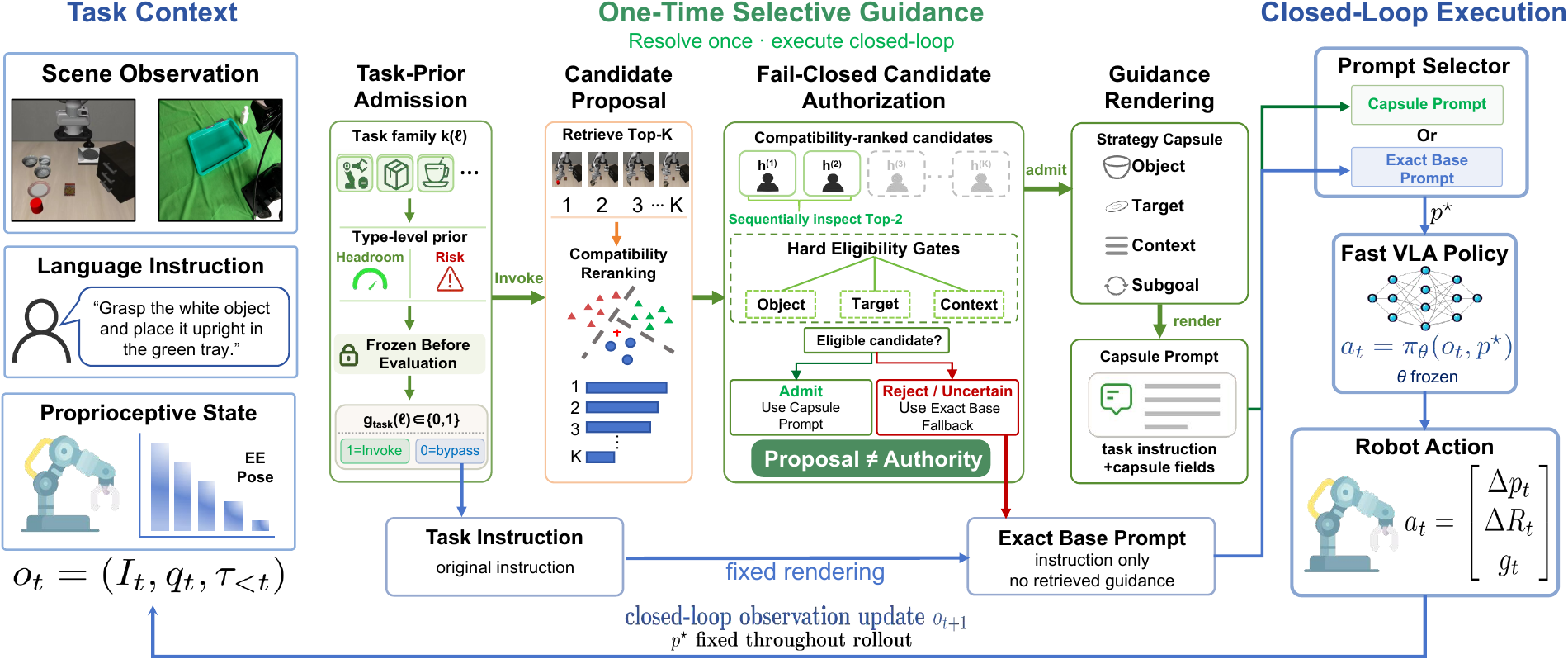}
\caption{TOWN-VLA separates invocation, candidate ranking,
prompt authorization, and frozen control. Ranked contexts are checked and
either rendered as a canonical capsule or rejected for exact Base restoration.
Logged scores, flags, and prompt hashes make routes auditable. Task-Prior is
used only for the oracle control in Equation~\eqref{eq:task-prior}.}
\label{fig:arch}
\end{figure*}

\section{Related Work}

\paragraph{Frozen policies and inference-time interfaces.}
PaLM-E, RT-1, and RT-2 established scalable language-conditioned robot control
\cite{driess2023palme,brohan2022rt1,brohan2023rt2}, while Open X-Embodiment,
Octo, and OpenVLA scaled cross-dataset adaptation
\cite{openx2023,octo2024,kim2024openvla}. Continuous and
latent-action models extend bimanual, spatial, and open-world control
\cite{liu2024rdt,black2024pi0,black2025pi05},
with deployability explored by SmolVLA
\cite{shukor2025smolvla}. These lines redesign the policy, training
data, or action representation. We instead freeze the generator and regulate
external text through the interface it already understands.

\paragraph{External reasoning and fast--slow authority.}
SayCan and VoxPoser ground language through affordances and spatial value maps
\cite{ahn2022saycan,huang2023voxposer}. ECoT,
InstructVLA, Acting While Understanding, and VLS move reasoning closer to action
\cite{zawalski2025ecot,yang2025instructvla,yan2026acting,liu2026vls}.
Fast--slow systems separate semantic and reactive computation
\cite{zhang2024hirt,chen2025fastinslow,zou2025duocore}.
Our slow path remains external: it may compute a proposal, but its text cannot
condition the action generator without authorization. The separation therefore
concerns control rights, not only execution rate.

\paragraph{Inspectable memory at the policy boundary.}
Unlike latent memory, an external capsule can be inspected and hashed before it
crosses the policy boundary. MemoryVLA and MemoryVLA++ integrate perceptual and
episodic history \cite{shi2026memoryvla,shi2026memoryvlaplus}.
WorldVLA, VLA-JEPA, WAMs, and latent-action world models instead
use predicted visual or latent dynamics
\cite{cen2025worldvla,sun2026vlajepa,wang2026wamsurvey,garrido2026latentaction}.
External memories retrieve experience or action priors
\cite{sridhar2025memer,liu2026goal2skill,lin2026worldpilot,li2025mapvla,zhang2026harnessvla}.
We instead ask when an inspectable textual proposal may replace the frozen
policy instruction.

\paragraph{Prompt brittleness beyond language generation.}
Language models can be insensitive to prompt meaning yet highly sensitive to
formatting or optimized suffixes
\cite{webson2022prompts,sclar2024promptformat,zou2023universal}. These studies
concern output quality in open-loop generation. In a frozen closed-loop VLA,
out-of-template text conditions actions, turning prompt brittleness into control
failure. We therefore regulate boundary crossing rather than optimize a prompt.

\paragraph{Selective intervention and runtime assurance.}
LIBERO-Plus, Q-DIG, and STRONG-VLA expose or train against visual and linguistic
shift \cite{fei2026liberoplus,srikanth2026qdig,xie2026strongvla}. Assistance
and routing methods calibrate uncertainty or select experts
\cite{ren2023askhelp,zhang2026corevla,ren2026routervla}; Mostly Harmless VLA
Steering learns when feedback may help \cite{jeong2026mostlyharmless}.
Runtime-assurance systems supervise learned controllers through Simplex or
action shielding \cite{seto1998simplex,alshiekh2018shielding}. Our action
generator and fallback remain fixed, so metareasoning or rejection
\cite{russell1991metareasoning,geifman2019selectivenet} becomes exact Base-prompt
restoration rather than controller switching.

\section{Method}
\label{sec:method}

\subsection{Problem Setup and Interface Contract}

\begin{center}
\small
\setlength{\tabcolsep}{3pt}
\begin{tabular}{@{}lp{0.76\columnwidth}@{}}
\toprule
Symbol & Meaning \\
\midrule
$\ell,\ u^\star,\ p^\star$ & Base instruction, resolved instruction, and
resolved prompt $\mathcal{P}(u^\star)$ \\
$h,\ \mathcal{H}_K$ & One memory candidate and the retrieved top-$K$ set \\
$g,\ j^\star$ & Slow-path invocation bit and authorized candidate rank \\
$\operatorname{sig}$ & Normalized object--relation--target task signature \\
\bottomrule
\end{tabular}
\end{center}

TOWN-VLA wraps a frozen VLA policy $\pi_\theta$, with $\theta$ fixed
throughout evaluation. It implements a deterministic interface contract that
constrains candidate-generated policy inputs while retaining $\pi_\theta$ as
the sole action generator. At step $t$, the policy receives
$o_t=(I_t,q_t,\tau_{<t})$, comprising the image, robot state, and action
history. A fixed renderer $\mathcal{P}$ maps the original instruction to the
Base prompt:
\begin{equation}
p_{\rm base}=\mathcal{P}(\ell),\qquad
a_t^{\rm base}=\pi_\theta(o_t,p_{\rm base}),\qquad
\theta\ \text{fixed}.
\label{eq:interface-contract}
\end{equation}
The interface initializes $u^\star=\ell$; only the authorization rule in
Equation~\eqref{eq:prompt-authority} may replace it with a canonical
instruction and set $p^\star=\mathcal{P}(u^\star)$. Otherwise
$p^\star=p_{\rm base}$ byte for byte. One resolved prompt is reused throughout
rollout, leaving policy parameters, action space, and control frequency fixed.

Retrieval logs candidate identifiers, reranking logs scores, the checker logs
its flag and reason, and the interface logs the resolved-prompt hash. Together
they separate proposal order, authorization, and controller input.

\subsection{Compatibility-Reranked Capsule}

The frozen memory $\mathcal{M}=\{h_i\}_{i=1}^{48}$ contains demonstration
trajectories but no evaluation rollouts. CLIP text features and a
nearest-neighbor index retrieve $K=5$ candidates
\cite{radford2021clip,johnson2019faiss}:
\begin{equation}
\mathcal{H}_K(\ell)=\operatorname{Retrieve}_K(\ell;\mathcal{M}).
\label{eq:retrieve}
\end{equation}
Each memory entry stores a task description, structured context, raw plan, and
compact hint. Retrieval indexes the description; parsing and rendering consume
only the context. Raw plans and hints remain provenance fields, never executed
text, and the memory is read-only during evaluation.

A frozen parser extracts object--target pairs from the task and each stored
context. Both parsers read text only---never images, robot state, action
history, or benchmark labels. Fixed pick/place templates and deterministic
fallbacks cover unmatched strings. With $J$ denoting token-set Jaccard overlap,
\begin{equation}
\begin{aligned}
x_\ell&=\operatorname{ParseTask}(\ell)
      =(x_{\rm obj},x_{\rm tgt}),\\
y_h&=\operatorname{ParseContext}(h)
      =(y_{\rm obj},y_{\rm tgt}),\\
m_{\rm obj}&=J(x_{\rm obj},y_{\rm obj}),\quad
m_{\rm tgt}=J(x_{\rm tgt},y_{\rm tgt}),\\
m_{\rm ctx}&=J(x_{\rm obj}\texttt{->}x_{\rm tgt},
                    \operatorname{context}(h)).
\end{aligned}
\label{eq:parsed-overlap}
\end{equation}
Here $x_{\rm obj}\texttt{->}x_{\rm tgt}$ is the literal ordered string formed
by concatenating the normalized object phrase, the delimiter
\texttt{->}, and the normalized target phrase; it is not a functional map.
For route identity, the same frozen parser supplies normalized object,
relation, and target fields:
\[
\begin{aligned}
\operatorname{sig}(u)
&=\bigl(c_{\rm obj}(u),c_{\rm rel}(u),c_{\rm tgt}(u)\bigr),\\
\operatorname{SigEq}(u,v)
&=\mathbf{1}\!\left[\operatorname{sig}(u)=\operatorname{sig}(v)\right].
\end{aligned}
\]
Signature equality is componentwise after the parser's deterministic text
normalization and fixed relation-alias mapping.
The mismatch indicators are
$r_{\rm obj}=\mathbf{1}[y_{\rm obj}\neq\varnothing\land m_{\rm obj}=0]$
and
$r_{\rm tgt}=\mathbf{1}[y_{\rm tgt}\neq\varnothing\land m_{\rm tgt}=0]$.
The frozen compatibility score is
\begin{equation}
\begin{aligned}
s(h,x_\ell)={}&
\alpha_{\rm clip}s_{\rm clip}(h,\ell)
 +\alpha_{\rm obj}m_{\rm obj}
 +\alpha_{\rm tgt}m_{\rm tgt}\\
&+\alpha_{\rm ctx}m_{\rm ctx}
-\lambda_{\rm obj}r_{\rm obj}
-\lambda_{\rm tgt}r_{\rm tgt}\\
&-\eta\,\operatorname{rank}_{\rm clip}(h).
\end{aligned}
\label{eq:compat-score}
\end{equation}
We fix $(\alpha_{\rm clip},\alpha_{\rm obj},\alpha_{\rm tgt},
\alpha_{\rm ctx},\lambda_{\rm obj},\lambda_{\rm tgt},\eta)
=(1,2,1.5,0.8,0.6,0.4,0.01)$ before evaluation. These coefficients rerank
textual agreement and are not fitted to outcomes. The resulting order is
\begin{equation}
(h_{(1)},\ldots,h_{(K)})
=\operatorname*{argsort}^{\downarrow}_{h\in\mathcal{H}_K(\ell)}
s(h,x_\ell).
\label{eq:ranked-candidates}
\end{equation}
The score combines CLIP proximity with structural overlap, explicit mismatch
penalties, and deterministic rank-based tie resolution. Reranking changes only
inspection order; the hard checker alone grants authorization.

\subsection{Top-2 Fail-Closed Cascade}

The hard checker records deterministic conflict reasons
$\mathcal{R}(h,\ell)$. Candidate eligibility is
\begin{equation}
G_{\rm comp}(h,\ell)
=\mathbf{1}\!\left[\mathcal{R}(h,\ell)=\varnothing\right].
\label{eq:compat-gate}
\end{equation}
The checker flags a nonempty candidate object or target with zero task overlap.
Context affects only soft ranking, so this remains a text-level rule. Rank 2 is
inspected only after rank 1 is rejected:
\begin{equation}
j^\star
=\min\!\left\{j\in\{1,2\}:
G_{\rm comp}(h_{(j)},\ell)=1\right\},
\label{eq:top2-cascade}
\end{equation}
with $j^\star=\varnothing$ if neither passes. Let $g\in\{0,1\}$ denote whether
the slow path is invoked; always-on evaluation sets $g=1$, while the oracle
ablation sets $g=g_{\rm prior}^{(m)}(z)$. The deterministic renderer and final
authority rule are
\begin{equation}
\begin{aligned}
u_{\rm cap}
&=\operatorname{RenderCapsule}\!\left(
\ell,\operatorname{context}(h_{(j^\star)})\right),\\
u^\star
&=
\begin{cases}
u_{\rm cap},&g=1\land j^\star\neq\varnothing,\\
\ell,&\text{otherwise},
\end{cases}\\
p^\star&=\mathcal{P}(u^\star).
\end{aligned}
\label{eq:prompt-authority}
\end{equation}
The cascade short-circuits after the first eligible candidate and never merges
contexts. An accepted context is rendered with the fixed template
\texttt{put <object> <relation> <target>}; empty or unrenderable context returns
$\ell$. For example, if Base is \texttt{Place the black bowl on the plate.},
rejection returns that byte-identical string, whereas authorization resolves to
\texttt{put the black bowl on the plate}. Fallback adds no retrieved text or rejection message. In the 1,200-state
audit, all authorized routes resolve at rank 1, leaving the second slot
unexercised. Top-2 therefore serves as a fixed safety budget that preserves
one recovery opportunity under future memory or ranking changes.

\subsection{Task-Prior Admission as an Oracle Control}

To quantify the oracle-side upper bound on removable slow-path computation,
a benchmark label $z$ supplies a manifest-indexed routing bit:
\begin{equation}
g_{\rm prior}^{(m)}(z)
=\mathbf{1}\!\left[z\in\mathcal{Z}_{\rm allow}^{(m)}\right],
\label{eq:task-prior}
\end{equation}
where $m\in\{\text{five-suite},\text{paired-3000}\}$ indexes the frozen
routing manifest.
For the five-suite manifest, $\mathcal{Z}_{\rm allow}^{(\text{five-suite})}$
contains red-stick, mug, and yellow-book spatial suites; standard-spatial and
object-mug execute Base. Its 3/5 allow--2/5 skip lookup saves 40\% of calls.
Table~\ref{tab:config}b instead uses a separately frozen paired-3000 manifest
with 1{,}500 invoked and 1{,}500 bypassed states, hence 50\% savings. The two
denominators are reported separately and never pooled. For $N$ episodes,
slow-path coverage and calls saved are
\begin{equation}
\rho_{\rm slow}
=\frac{1}{N}\sum_{i=1}^{N}g_i,\qquad
S_{\rm call}=1-\rho_{\rm slow}.
\label{eq:route-use}
\end{equation}
Routing precedes retrieval and changes only whether slow-path work runs;
Equations~\eqref{eq:compat-gate}--\eqref{eq:prompt-authority} separately control
prompt intervention. Per episode, the interface adds at most one retrieval,
five scores, two checks, one rendering, and one prompt resolution. All
components and routing tables are frozen before evaluation.

\subsection{Runtime States and Interface Guarantees}

TOWN-VLA exposes three auditable runtime states: bypass executes $\mathcal{P}(\ell)$
without retrieval; inspected-but-unauthorized also executes Base after checking;
authorized renders $u_{\rm cap}$. Thus spending slow-path computation and
granting prompt authority remain separate decisions.

The construction enforces (I1) prompt-boundary non-interference on rejection,
(I2) $\pi_\theta$ as the sole action generator, and (I3) a memory-size-independent
post-retrieval inspection budget. Candidate ID, conflict reason, route state,
and resolved-prompt hash provide the corresponding execution trace.

Execution initializes $u^\star=\ell$, evaluates routing, and, if invoked,
retrieves, scores, and checks at most two candidates. One prompt hash is fixed
before the first action; the slow path is not queried again during rollout.

Unless labeled oracle-side, all results use $g=1$; the main evaluation
therefore measures prompt authority independently of benchmark labels and
Task-Prior routing.

\begin{table*}[!t]
\centering
\small
\setlength{\tabcolsep}{4pt}
\begin{tabular}{llccccccc c}
\toprule
Method & Setting & Camera & Robot & Language & Light & Background & Noise & Layout & Weighted Mean \\
\midrule
OpenVLA           & Direct                  & 0.8  & 3.5  & 23.0 & 8.1  & 34.8 & 15.2 & 28.5 & 15.6 \\
NORA              & Direct                  & 2.2  & 37.0 & 65.1 & 45.7 & 58.6 & 12.8 & 62.1 & 39.0 \\
WorldVLA          & Direct                  & 0.1  & 27.9 & 41.6 & 43.7 & 17.1 & 10.9 & 38.0 & 25.0 \\
UniVLA            & Direct                  & 1.8  & 46.2 & 69.6 & 69.0 & 81.0 & 21.2 & 31.9 & 42.9 \\
\pipizero{}       & Flow policy             & 13.8 & 6.0  & 58.8 & 85.0 & 81.4 & 79.0 & 68.9 & 53.6 \\
\pipizero{}-FAST  & Autoregressive          & 65.1 & 21.6 & 61.0 & 73.2 & 73.2 & 74.4 & 68.8 & 61.6 \\
OpenVLA-OFT$_w$   & Third-view only         & 10.4 & 38.7 & 70.5 & 76.8 & 93.6 & 49.9 & 69.9 & 55.8 \\
OpenVLA-OFT$_m$   & Mixed supervised tuning & 55.6 & 21.7 & 81.0 & 92.7 & 91.0 & 78.6 & 68.7 & 67.9 \\
OpenVLA-OFT       & Optimized fine-tuning     & 56.4 & 31.5 & 78.7 & 88.6 & 93.1 & 75.6 & 74.9 & 69.5 \\
TOWN-VLA (Always-On)& Frozen-backbone interface & 61.5 & 37.5 & 85.2 & 92.7 & 95.6 & 73.6 & 78.0 & 73.1 \\
\bottomrule
\end{tabular}
\caption{Seven-axis LIBERO-Plus SR (\%). Published rows are
from~\citet{fei2026liberoplus}; the local OpenVLA-OFT and TOWN-VLA rows use the same frozen backbone,
28 cells, and 10,030 episodes per method.}
\label{tab:liberoplus}
\end{table*}

\section{Experiments}

We organize the evaluation around five questions: end-to-end robustness (Q1),
prompt-form failure (Q2), authorization and exact restoration (Q3), selective
computation (Q4), and transfer across backbones and embodiments (Q5).

\subsection{Setup and Protocol Map}

OpenVLA with the Optimized Fine-Tuning recipe (OpenVLA-OFT) is the Base
Policy~\cite{kim2025finetuning}. TOWN-VLA and its ablations modify only
the prompt interface while holding the backbone, action space, policy
parameters, and control frequency fixed.
Table 1 and all Q1--Q3/Q5 end-to-end comparisons use the always-on setting
($g=1$), independently of benchmark labels and Task-Prior routing.
Experiments run on NVIDIA RTX A6000 hardware with CUDA 12.1 and headless MuJoCo
rendering through EGL~\cite{todorov2012mujoco}. Success rate (SR) is the
primary metric.

Each frozen protocol includes its own matched Base Policy, and every reported
difference uses the Base evaluated on the same states.
Simulation contrasts use paired initial states whenever episode-level outcomes
are available. The LIBERO-Plus interval resamples its 28 matched suite--axis
cells, the prompt-form audit uses paired bootstrap and exact McNemar tests, and
physical task success uses a two-sided Fisher exact test. Published comparison
rows provide context only; claims about the interface use the locally matched
backbone and execution stack. The LIBERO-Plus interval summarizes variation
across matched cells, rather than independent run-to-run variation.

\subsection{Perturbation Breadth (Q1)}

Under the matched local protocol, TOWN-VLA raises weighted LIBERO-Plus SR
by 3.61 points, adding 362 successes over 10{,}030 episodes against the same
OpenVLA-OFT backbone (Table~\ref{tab:liberoplus}). The paired-cell 95\%
interval is 1.89--5.45 points. The axis view improves in six of seven
conditions, led by Language ($+6.5$) and Robot ($+6.0$); image-only Noise is
the sole lower estimate. Aggregating the same episodes by task suite gives a
positive difference for Spatial, Object, Goal, and Long-Horizon
(Table~\ref{tab:four-suite}). The two partitions show that the overall gain is
distributed across perturbation types and task families.

\begin{center}
\begin{minipage}{\columnwidth}
\centering
\small
\setlength{\tabcolsep}{2.2pt}
\begin{tabular*}{\linewidth}{@{\extracolsep{\fill}}lccccc@{}}
\toprule
Method & Spatial & Object & Goal & Long & Overall \\
\midrule
OpenVLA           & 19.4 & 14.0 & 15.1 & 14.3 & 15.6 \\
WorldVLA          & 32.5 & 28.6 & 31.8 & 8.2  & 25.0 \\
UniVLA            & 55.5 & 36.7 & 40.7 & 39.9 & 42.9 \\
\pipizero{}       & 60.7 & 61.4 & 44.9 & 48.4 & 53.6 \\
\pipizero{}-FAST  & 74.4 & 72.7 & 57.6 & 43.4 & 61.6 \\
\midrule
OpenVLA-OFT       & 84.0 & 65.8 & 62.9 & 65.9 & 69.5 \\
TOWN-VLA& 86.3 & 69.5 & 68.1 & 69.1 & 73.1 \\
\bottomrule
\end{tabular*}
\captionof{table}{Four-suite LIBERO-Plus SR (\%). Published rows follow
\citet{fei2026liberoplus}; local rows aggregate the matched episodes in
Table~\ref{tab:liberoplus}.}
\label{tab:four-suite}
\end{minipage}
\end{center}

\subsection{Prompt-Form Collapse (Q2)}

\begin{figure}[!t]
\centering
\includegraphics[width=\columnwidth]{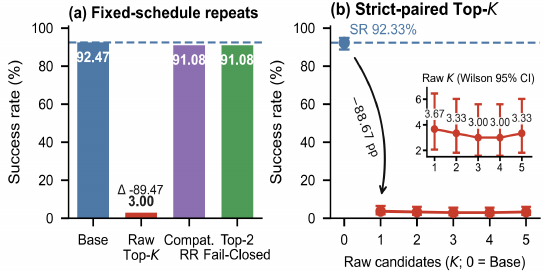}
\caption{Prompt-form collapse and retrieval-depth diagnostics. (a) Mean SR
over three fixed-schedule repeats on one manifest (1{,}200 episodes per method
and repeat); the Raw annotation is its change from matched Base. Compat.\ RR
and Top-2 denote the reranked capsule and fail-closed cascade. (b) Paired
Top-$K$ results on 300 states with Wilson 95\% intervals.}
\label{fig:harm}
\end{figure}

Across three fixed-schedule executions, mean SR falls from 92.47\% under Base
to 3.00\% under raw appended text. Compatibility reranking and Top-2
fail-closed both recover to 91.08\% (Figure~\ref{fig:harm}a). The executions
reuse all 1{,}200 initial-state hashes and measure implementation
reproducibility, not environmental-seed variance. In the paired diagnostic,
the first raw candidate changes SR by $-88.67$ points (95\% CI
$-94.00$ to $-82.33$). Increasing $K$ beyond one neither worsens nor mitigates
the collapse, so retrieval depth is not the operative variable.

A matched 500-state factorial control separates prompt form from semantics
(Table~\ref{tab:form-semantics}). Base and exact restoration each reach
499/500, and the correct canonical instruction reaches 497/500. Correct and
meaningless appends both yield 0/500, whereas canonical wrong-object and
wrong-target controls reach 497/500 and 496/500. Task-only and retrieved
canonical arms share all prompt hashes and outcomes, indicating that prompt
form, rather than retrieved strategy content, explains performance in this
audit.

\begin{center}
\begin{minipage}{\columnwidth}
\centering
\small
\setlength{\tabcolsep}{3pt}
\begin{tabular*}{\linewidth}{@{\extracolsep{\fill}}lcccc@{}}
\toprule
Prompt form & Correct & Wrong obj. & Wrong tgt. & Meaningless \\
\midrule
Canonical & 497/500 & 497/500 & 496/500 & \na \\
Appended  & 0/500   & \na     & \na     & 0/500 \\
\bottomrule
\end{tabular*}
\captionof{table}{Prompt form $\times$ semantics on 500 matched states.
Canonical controls alter the object or target; the meaningless append is
length matched. Base and exact restoration are 499/500.}
\label{tab:form-semantics}
\end{minipage}
\end{center}

\begin{table*}[!t]
\centering
\small
\begin{minipage}[t]{0.60\textwidth}
\centering
(a) Interface configuration, five-suite manifest\\[2pt]
\resizebox{\linewidth}{!}{%
\begin{tabular}{lccccc}
\toprule
Variant & Task-prior & Capsule & Rerank & Fail-closed & SR (\%) \\
\midrule
Base Policy                    & \na & \na & \na & \na & 89.48 \\
Strategy Capsule (Top-1)       & \na & \ck & \na & \na & 90.64 \\
Compatibility-Reranked Capsule & \na & \ck & \ck & \na & 91.60 \\
Top-2 Fail-Closed Cascade      & \na & \ck & \ck & \ck & 91.32 \\
Complete Interface             & \ck & \ck & \ck & \ck & 91.32 \\
\bottomrule
\end{tabular}
}
\end{minipage}
\hfill
\begin{minipage}[t]{0.38\textwidth}
\centering
(b) Oracle-side selective invocation\\[2pt]
\resizebox{\linewidth}{!}{%
\begin{tabular}{lccc}
\toprule
Metric & Always-On & Routed & Change \\
\midrule
Paired SR (\%)       & 94.20  & 94.20  & $0.00$ \\
Slow-path calls      & 3000/3000 & 1500/3000 & $-50\%$ \\
Decision latency (ms)& 212.50 & 190.75 & $-10.24\%$ \\
\bottomrule
\end{tabular}
}
\end{minipage}
\caption{Interface and invocation controls. Complete Interface is TOWN-VLA. (a) Five-suite SR over 2{,}500
episodes per variant; checks mark enabled components. (b) Oracle routing on a
separate paired-3{,}000 manifest, split equally between invoke and bypass.}
\label{tab:config}
\end{table*}

\subsection{Authorization and Exact Restoration (Q3)}

\paragraph{Route-level identity.}
A 1{,}200-state always-on audit separates authorization from restoration.
Reranking admits all states and preserves the task signature on 1{,}100/1{,}200.
The cascade admits 1{,}000 states, preserves the signature on 940/1{,}000
authorized routes, and restores exact Base on the other 200. Thus its 94.00\%
denominator is the authorized subset, not all states.

A separate 900-route audit includes routing. The slow path runs on 450 routes:
375 are authorized with $\operatorname{SigEq}=1$, and 75 are inspected then
rejected. The other 450 bypass retrieval. Consequently, the 525 exact-Base
routes comprise 450 bypasses and 75 rejections. The identity and restoration
rates differ because the two audits use different manifests and denominators.
The high-Base always-on audit yields +1.00/+1.67 points for
signature-changing prompts (McNemar $p=1$), whereas Q1 localizes gains to
perturbed inputs, especially the lower-Base Language and Robot cells. This
contrast is consistent with task-equivalent canonicalization rather than
semantic departure.

\paragraph{Restricted-OOD replication.}
The same ordering appears under spatial/object shift
(Table~\ref{tab:ood}): Raw Top-5 drops to 74.20/79.40, reranking recovers to
88.20/96.67, and fail-closed authorization reaches 89.33/97.53, above Base in
both reported shift aggregates. We additionally evaluate five released
checkpoints under the same local protocol. The citations identify the model
sources rather than supply these measurements
\cite{sun2026vlajepa,lerobot2026vlajepalibero,shukor2025smolvla,lerobot2025smolvlalibero,beeface2026smolvlaliberospatial};
their locally measured spatial means span 55.20\%--88.13\%, and object means
span 45.60\%--95.07\%.
On the five-suite manifest (Table~\ref{tab:config}a), reranking scores 91.60\%
and the bounded cascade 91.32\%. The 0.28-point difference is seven outcomes
over 2{,}500 episodes; the cascade is retained for exact restoration rather
than claimed as an incremental SR mechanism.

\begin{center}
\begin{minipage}{\columnwidth}
\centering
\small
\setlength{\tabcolsep}{4pt}
\begin{tabular}{lcc}
\toprule
Method & Spatial shift & Object shift \\
\midrule
Base Policy                     & 85.80 & 93.93 \\
Raw Top-5 Plan                  & 74.20 & 79.40 \\
Compatibility-Reranked Capsule  & 88.20 & 96.67 \\
Top-2 Fail-Closed Cascade       & 89.33 & 97.53 \\
\bottomrule
\end{tabular}
\captionof{table}{Restricted-OOD success rate (\%) under spatial and object
shifts. All rows are local evaluations sharing one frozen Base Policy and
execution protocol.}
\label{tab:ood}
\end{minipage}
\end{center}

\subsection{Selective Computation and Gate Controls (Q4)}
\label{sec:q4}

\paragraph{Oracle-side headroom.}
On 3{,}000 paired states, benchmark-side routing preserves the same 2{,}826
successes while halving slow-path calls (Table~\ref{tab:config}b). Decision
latency falls 10.24\%, from 212.50 to 190.75\,ms, over three rounds of 300
decisions after 50 warm-ups. This episode-level decision includes routing,
retrieval, reranking, tokenization, and one VLA decode, but excludes simulator
and startup time. The five-suite manifest separately saves 40\% of calls.

\paragraph{Oracle-free admission remains a calibration challenge. }
On the preregistered 24-development/36-held-out cell split, the learned selector authorizes 2/36 cells and preserves the paired outcomes across those 40 episodes, matching the 91.81\% Base Policy; CLIP authorizes 0/36 cells. Always-on intervention is 1.53 points below Base, and the form-aware and matched-budget controls do not recover this gap. Thus, the oracle analysis quantifies removable computation, while the evaluated gates do not yet identify which calls to remove.

A separate compositional-language probe identifies the relational fields
needed to extend canonical rendering. Base reaches 20/30, whereas canonical
prompts without drawer, cabinet, and relative-location clauses reach 0/30.
Together with the 500-state object--target control, this result bounds the
current renderer and motivates preserving richer relational structure for
compositional instructions.

\begin{center}
\begin{minipage}{\columnwidth}
\centering
\scriptsize
\setlength{\tabcolsep}{3pt}
\begin{tabular*}{\linewidth}{@{\extracolsep{\fill}}lcc@{}}
\toprule
Arm & Authorized cells & SR (\%) \\
\midrule
Exact Base (reference) & 0/36 & 91.81 \\
Learned, no oracle & 2/36 & 91.81 \\
Fixed/random matched budget & 32/36 & 90.14--90.97 \\
Form-aware & 32/36 & 90.28 \\
\bottomrule
\end{tabular*}
\captionof{table}{Oracle-free gates on the fixed 24-development/36-held-out
split, reporting held-out authorization and SR.}
\label{tab:gates-compact}
\end{minipage}
\end{center}

\begin{figure*}[!t]
\centering
\includegraphics[width=0.92\textwidth]{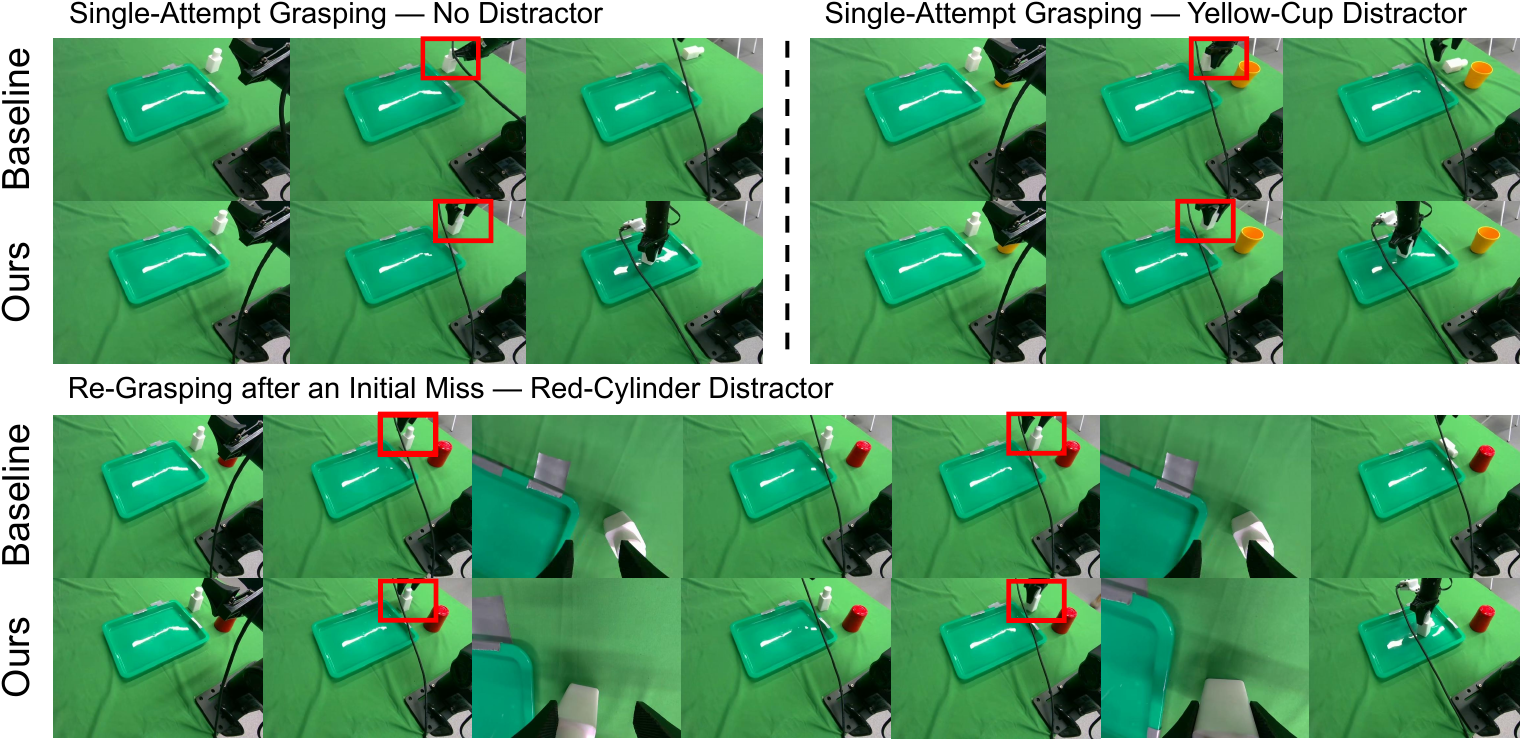}
\caption{Representative \pizerofive{}/PiPER rollouts in the three evaluated
scenes. Columns are time ordered; rows compare Base and TOWN-VLA, and red boxes
mark interaction regions. Prompts resolve once; results use all 150 trials per
method.}
\label{fig:piper}
\end{figure*}

\subsection{Transfer to a Second Backbone and a Physical Robot (Q5)}

The instruction is ``Grasp the white object and place it upright in the green
tray.'' We test no-distractor, yellow-cup, and red-cylinder scenes. Object poses
are randomized around a teleoperated reference, and Base/TOWN-VLA trials are
randomly interleaved. A human records success after the target remains upright
for two seconds (50 trials per scene--method cell).

Both methods share the PiPER arm, dual RealSense D405 cameras, frozen
\pizerofive{} checkpoint, and success criterion. Controller settings are fixed
to a 10-action-step replanning horizon, speed ratio 5, and a 60-second limit.
Drops, collisions, human intervention, timeout, or failure to maintain the
upright placement count as failures. Formal trials are excluded from training,
memory, and calibration, leaving the prompt interface as the method-level
change.

TOWN-VLA raises success from 79/150 to 118/150
($+26.00$ points; $p=3.16\times10^{-6}$, Fisher exact test;
Table~\ref{tab:piper}), with 22--32-point gains in all three scenes. Because
prompts resolve before rollout, corrective re-grasp is closed-loop policy
behavior (Figure~\ref{fig:piper}). Post-miss recovery rises from 24.4\% to
52.4\%, but remains descriptive: method-dependent misses create unequal
post-treatment denominators (41 versus 21).

Because pose distributions differ by scene, the controlled comparison is Base
versus TOWN-VLA (Complete Interface) within each row. All three favor the Complete Interface, so the pooled gain
is not carried by one configuration.

\begin{center}
\begin{minipage}{\columnwidth}
\centering
\scriptsize
\resizebox{\columnwidth}{!}{%
\begin{tabular}{lcc cc}
\toprule
& \multicolumn{2}{c}{Task success} & \multicolumn{2}{c}{Post-miss recovery} \\
\cmidrule(lr){2-3}\cmidrule(lr){4-5}
Scene & Base Policy & TOWN-VLA & Base Policy & TOWN-VLA \\
\midrule
No distractor     & 21/50 (42.0) & 37/50 (74.0)  & 3/16 (18.8)  & 2/6 (33.3)  \\
Yellow distractor & 27/50 (54.0) & 39/50 (78.0)  & 3/13 (23.1)  & 3/7 (42.9)  \\
Red distractor    & 31/50 (62.0) & 42/50 (84.0)  & 4/12 (33.3)  & 6/8 (75.0)  \\
\midrule
Overall           & 79/150 (52.7)& 118/150 (78.7)& 10/41 (24.4) & 11/21 (52.4)\\
\bottomrule
\end{tabular}
}
\captionof{table}{\pizerofive{}/PiPER task success and descriptive recovery.
Cells show successes/trials (\%); recovery conditions on an initial miss and
uses method-dependent denominators.}
\label{tab:piper}
\end{minipage}
\end{center}

These trials evaluate interface transfer rather than policy adaptation:
backbone, cameras, feedback loop, and criterion remain fixed; only the resolved
instruction changes.

\section{Discussion}

TOWN-VLA results support separating retrieval relevance from prompt authority: canonical rendering constrains accepted intervention, while exact restoration makes rejection reversible. Across protocols, the contrast between Q1 and Q3 suggests that intervention is most valuable when perturbations have degraded the executed instruction and is largely neutral when Base remains strong. Route-level attribution of the Q1 gain is an important next step. Simulation isolates this interface behavior under matched manifests, while PiPER tests the same behavior on a second backbone and embodiment. The present evaluation deliberately isolates the interface using a 48-entry same-domain memory, text-only compatibility, and a controlled single-task, single-operator physical study with human-scored outcomes. Natural extensions include larger cross-domain memories, visually conditioned admission, and broader blinded robot trials.

\section{Conclusion}

We studied how retrieved text should cross the input boundary of a frozen
VLA. Prompt-form controls identify severe prompt-form collapse under raw
appends, while TOWN-VLA (Think Only When Needed) separates candidate
generation from authorization and restores Base exactly on rejected routes.
Under matched evaluation, the interface improves LIBERO-Plus by 3.61 points
and the PiPER real-robot platform by 26.00 points without retraining the
action generator. These results establish prompt authority as an enforceable
control primitive for frozen VLAs and identify reliable oracle-free admission
as the next frontier for selective slow-path control.

\clearpage
{\small
\bibliography{references}

@inproceedings{brohan2023rt2,
  title = {{RT-2}: Vision-Language-Action Models Transfer Web Knowledge to Robotic Control},
  author = {Zitkovich, Brianna and others},
  booktitle = {Proceedings of The 7th Conference on Robot Learning},
  pages = {2165--2183},
  volume = {229},
  series = {Proceedings of Machine Learning Research},
  publisher = {PMLR},
  year = {2023},
  url = {https://proceedings.mlr.press/v229/zitkovich23a.html}
}

@inproceedings{kim2024openvla,
  title = {{OpenVLA}: An Open-Source Vision-Language-Action Model},
  author = {Kim, Moo Jin and Pertsch, Karl and Karamcheti, Siddharth and Xiao, Ted and Balakrishna, Ashwin and Nair, Suraj and Rafailov, Rafael and Foster, Ethan P. and Sanketi, Pannag R. and Vuong, Quan and Kollar, Thomas and Burchfiel, Benjamin and Tedrake, Russ and Sadigh, Dorsa and Levine, Sergey and Liang, Percy and Finn, Chelsea},
  booktitle = {Proceedings of The 8th Conference on Robot Learning},
  pages = {2679--2713},
  volume = {270},
  series = {Proceedings of Machine Learning Research},
  publisher = {PMLR},
  year = {2025},
  url = {https://proceedings.mlr.press/v270/kim25c.html}
}

@inproceedings{zawalski2025ecot,
  title = {Robotic Control via Embodied Chain-of-Thought Reasoning},
  author = {Zawalski, Micha{\l} and Chen, William and Pertsch, Karl and Mees, Oier and Finn, Chelsea and Levine, Sergey},
  booktitle = {Proceedings of The 8th Conference on Robot Learning},
  pages = {3157--3181},
  volume = {270},
  series = {Proceedings of Machine Learning Research},
  publisher = {PMLR},
  year = {2025},
  url = {https://proceedings.mlr.press/v270/zawalski25a.html}
}

@inproceedings{zhang2024hirt,
  title = {{HiRT}: Enhancing Robotic Control with Hierarchical Robot Transformers},
  author = {Zhang, Jianke and Guo, Yanjiang and Chen, Xiaoyu and Wang, Yen-Jen and Hu, Yucheng and Shi, Chengming and Chen, Jianyu},
  booktitle = {Proceedings of The 8th Conference on Robot Learning},
  pages = {933--946},
  volume = {270},
  series = {Proceedings of Machine Learning Research},
  publisher = {PMLR},
  year = {2025},
  url = {https://proceedings.mlr.press/v270/zhang25b.html}
}

@misc{li2025mapvla,
  title = {{MAP-VLA}: Memory-Augmented Prompting for Vision-Language-Action Model in Robotic Manipulation},
  author = {Li, Runhao and Guo, Wenkai and Wu, Zhenyu and Wang, Changyuan and Deng, Haoyuan and Weng, Zhenyu and Tan, Yap-Peng and Wang, Ziwei},
  year = {2025},
  eprint = {2511.09516},
  archivePrefix = {arXiv},
  primaryClass = {cs.RO},
  doi = {10.48550/arXiv.2511.09516},
  url = {https://arxiv.org/abs/2511.09516}
}

@misc{yan2026acting,
  title = {Acting While Understanding: Asynchronous Semantic-Action Decoupling for Real-Time Vision-Language-Action Models},
  author = {Yan, Shenhao and Wang, Ge and Liu, Qi and Meng, Weilin and Yang, Jiahao and Yao, Chengsi and Feng, Fan and Ma, Xiaoguang and Zhao, Yiming and Han, Yatong},
  year = {2026},
  eprint = {2606.15285},
  archivePrefix = {arXiv},
  primaryClass = {cs.RO},
  doi = {10.48550/arXiv.2606.15285},
  url = {https://arxiv.org/abs/2606.15285}
}

@inproceedings{ren2023askhelp,
  title = {Robots That Ask For Help: Uncertainty Alignment for Large Language Model Planners},
  author = {Ren, Allen Z. and Dixit, Anushri and Bodrova, Alexandra and Singh, Sumeet and Tu, Stephen and Brown, Noah and Xu, Peng and Takayama, Leila and Xia, Fei and Varley, Jake and Xu, Zhenjia and Sadigh, Dorsa and Zeng, Andy and Majumdar, Anirudha},
  booktitle = {Proceedings of The 7th Conference on Robot Learning},
  pages = {661--682},
  volume = {229},
  series = {Proceedings of Machine Learning Research},
  publisher = {PMLR},
  year = {2023},
  url = {https://proceedings.mlr.press/v229/ren23a.html}
}

@misc{zhang2026corevla,
  title = {{CoRE-VLA}: Towards Scalable and Robust Vision-Language-Action Modeling via Conditional Routing of Experts},
  author = {Zhang, Haozhe and Li, Sixian and Zhang, Yifei and Huai, Zezheng and Chen, Hao and Shen, Chunhua and Gong, Jingjing and Qiu, Xipeng},
  year = {2026},
  eprint = {2607.03693},
  archivePrefix = {arXiv},
  primaryClass = {cs.RO},
  doi = {10.48550/arXiv.2607.03693},
  url = {https://arxiv.org/abs/2607.03693}
}

@misc{ren2026routervla,
  title = {{RouterVLA}: Budgeted Commissioning and Expert Onboarding for Growing {VLA} Pools},
  author = {Ren, Xingyu and Yi, Chugang and Sun, Youran},
  year = {2026},
  eprint = {2606.27355},
  archivePrefix = {arXiv},
  primaryClass = {cs.RO},
  doi = {10.48550/arXiv.2606.27355},
  url = {https://arxiv.org/abs/2606.27355}
}

@misc{zhang2026harnessvla,
  title = {{Harness VLA}: Steering Frozen {VLA}s into Reliable Manipulation Primitives via Memory-Guided Agents},
  author = {Zhang, Yixian and others},
  year = {2026},
  eprint = {2607.08448},
  archivePrefix = {arXiv},
  primaryClass = {cs.RO},
  doi = {10.48550/arXiv.2607.08448},
  url = {https://arxiv.org/abs/2607.08448}
}

@misc{jeong2026mostlyharmless,
  title = {Learning What to Say to Your {VLA}: Mostly Harmless Vision Language Action Model Steering},
  author = {Jeong, Hyun Joe and Swamy, Gokul and Bajcsy, Andrea},
  year = {2026},
  eprint = {2606.12299},
  archivePrefix = {arXiv},
  primaryClass = {cs.RO},
  doi = {10.48550/arXiv.2606.12299},
  url = {https://arxiv.org/abs/2606.12299}
}

@inproceedings{fei2026liberoplus,
  title = {{LIBERO-Plus}: A Progressive Robustness Benchmark for Visual-Language-Action Models},
  author = {Fei, Senyu and Wang, Siyin and Shi, Junhao and Dai, Zihao and Cai, Jikun and Qian, Pengfang and Ji, Li and He, Xinzhe and Zhang, Shiduo and Fei, Zhaoye and Fu, Jinlan and Gong, Jingjing and Qiu, Xipeng},
  booktitle = {Proceedings of the IEEE/CVF Conference on Computer Vision and Pattern Recognition (CVPR)},
  pages = {38574--38583},
  month = {June},
  year = {2026},
  url = {https://openaccess.thecvf.com/content/CVPR2026/html/Fei_LIBERO-Plus_A_Progressive_Robustness_Benchmark_for_Visual-Language-Action_Models_CVPR_2026_paper.html}
}

@inproceedings{black2025pi05,
  title = {{$\pi_{0.5}$}: a Vision-Language-Action Model with Open-World Generalization},
  author = {Black, Kevin and Brown, Noah and Darpinian, James and Dhabalia, Karan and Driess, Danny and Esmail, Adnan and Equi, Michael Robert and Finn, Chelsea and Fusai, Niccolo and Galliker, Manuel Y. and Ghosh, Dibya and Groom, Lachy and Hausman, Karol and Ichter, Brian and Jakubczak, Szymon and Jones, Tim and Ke, Liyiming and LeBlanc, Devin and Levine, Sergey and Li-Bell, Adrian and Mothukuri, Mohith and Nair, Suraj and Pertsch, Karl and Ren, Allen Z. and Shi, Lucy Xiaoyang and Smith, Laura and Springenberg, Jost Tobias and Stachowicz, Kyle and Tanner, James and Vuong, Quan and Walke, Homer and Walling, Anna and Wang, Haohuan and Yu, Lili and Zhilinsky, Ury},
  booktitle = {Proceedings of The 9th Conference on Robot Learning},
  pages = {17--40},
  volume = {305},
  series = {Proceedings of Machine Learning Research},
  publisher = {PMLR},
  year = {2025},
  url = {https://proceedings.mlr.press/v305/black25a.html}
}

@misc{sridhar2025memer,
  title = {{MemER}: Scaling Up Memory for Robot Control via Experience Retrieval},
  author = {Sridhar, Ajay and Pan, Jennifer and Sharma, Satvik and Finn, Chelsea},
  year = {2025},
  eprint = {2510.20328},
  archivePrefix = {arXiv},
  primaryClass = {cs.RO},
  doi = {10.48550/arXiv.2510.20328},
  url = {https://arxiv.org/abs/2510.20328}
}

@article{russell1991metareasoning,
  title = {Principles of Metareasoning},
  author = {Russell, Stuart and Wefald, Eric},
  journal = {Artificial Intelligence},
  volume = {49},
  number = {1--3},
  pages = {361--395},
  year = {1991},
  doi = {10.1016/0004-3702(91)90015-C}
}

@inproceedings{geifman2019selectivenet,
  title = {{SelectiveNet}: A Deep Neural Network with an Integrated Reject Option},
  author = {Geifman, Yonatan and El-Yaniv, Ran},
  booktitle = {Proceedings of the 36th International Conference on Machine Learning},
  pages = {2151--2159},
  volume = {97},
  series = {Proceedings of Machine Learning Research},
  publisher = {PMLR},
  year = {2019},
  url = {https://proceedings.mlr.press/v97/geifman19a.html}
}

@article{seto1998simplex,
  title = {Dynamic Control System Upgrade Using the Simplex Architecture},
  author = {Seto, Danbing and Krogh, Bruce H. and Sha, Lui and Chutinan, Anuradha},
  journal = {IEEE Control Systems Magazine},
  volume = {18},
  number = {4},
  pages = {72--80},
  month = {August},
  year = {1998},
  doi = {10.1109/37.710880}
}

@inproceedings{radford2021clip,
  title = {Learning Transferable Visual Models From Natural Language Supervision},
  author = {Radford, Alec and Kim, Jong Wook and Hallacy, Chris and Ramesh, Aditya and Goh, Gabriel and Agarwal, Sandhini and Sastry, Girish and Askell, Amanda and Mishkin, Pamela and Clark, Jack and Krueger, Gretchen and Sutskever, Ilya},
  booktitle = {Proceedings of the 38th International Conference on Machine Learning},
  pages = {8748--8763},
  volume = {139},
  series = {Proceedings of Machine Learning Research},
  publisher = {PMLR},
  year = {2021},
  url = {https://proceedings.mlr.press/v139/radford21a.html}
}

@article{johnson2019faiss,
  title = {Billion-Scale Similarity Search with {GPUs}},
  author = {Johnson, Jeff and Douze, Matthijs and J{\'e}gou, Herv{\'e}},
  journal = {IEEE Transactions on Big Data},
  volume = {7},
  number = {3},
  pages = {535--547},
  year = {2021},
  doi = {10.1109/TBDATA.2019.2921572}
}

@inproceedings{kim2025finetuning,
  title = {Fine-Tuning Vision-Language-Action Models: Optimizing Speed and Success},
  author = {Kim, Moo Jin and Finn, Chelsea and Liang, Percy},
  booktitle = {Proceedings of Robotics: Science and Systems},
  address = {Los Angeles, CA, USA},
  month = {June},
  year = {2025},
  doi = {10.15607/RSS.2025.XXI.017},
  url = {https://www.roboticsproceedings.org/rss21/p017.html}
}

@inproceedings{todorov2012mujoco,
  title = {{MuJoCo}: A Physics Engine for Model-Based Control},
  author = {Todorov, Emanuel and Erez, Tom and Tassa, Yuval},
  booktitle = {IEEE/RSJ International Conference on Intelligent Robots and Systems},
  pages = {5026--5033},
  year = {2012},
  doi = {10.1109/IROS.2012.6386109}
}

@inproceedings{black2024pi0,
  title = {{$\pi_0$}: A Vision-Language-Action Flow Model for General Robot Control},
  author = {Black, Kevin and Brown, Noah and Driess, Danny and Esmail, Adnan and Equi, Michael Robert and Finn, Chelsea and Fusai, Niccolo and Groom, Lachy and Hausman, Karol and Ichter, Brian and Jakubczak, Szymon and Jones, Tim and Ke, Liyiming and Levine, Sergey and Li-Bell, Adrian and Mothukuri, Mohith and Nair, Suraj and Pertsch, Karl and Shi, Lucy Xiaoyang and Smith, Laura and Tanner, James and Vuong, Quan and Walling, Anna and Wang, Haohuan and Zhilinsky, Ury},
  booktitle = {Proceedings of Robotics: Science and Systems},
  address = {Los Angeles, CA, USA},
  month = {June},
  year = {2025},
  doi = {10.15607/RSS.2025.XXI.010},
  url = {https://www.roboticsproceedings.org/rss21/p010.html}
}

@inproceedings{pertsch2025fast,
  title = {{FAST}: Efficient Action Tokenization for Vision-Language-Action Models},
  author = {Pertsch, Karl and Stachowicz, Kyle and Ichter, Brian and Driess, Danny and Nair, Suraj and Vuong, Quan and Mees, Oier and Finn, Chelsea and Levine, Sergey},
  booktitle = {Proceedings of Robotics: Science and Systems},
  address = {Los Angeles, CA, USA},
  month = {June},
  year = {2025},
  doi = {10.15607/RSS.2025.XXI.012},
  url = {https://www.roboticsproceedings.org/rss21/p012.html}
}

@misc{cen2025worldvla,
  title = {{WorldVLA}: Towards Autoregressive Action World Model},
  author = {Cen, Jun and Yu, Chaohui and Yuan, Hangjie and Jiang, Yuming and Huang, Siteng and Guo, Jiayan and Li, Xin and Song, Yibing and Luo, Hao and Wang, Fan and Zhao, Deli and Chen, Hao},
  year = {2025},
  eprint = {2506.21539},
  archivePrefix = {arXiv},
  primaryClass = {cs.RO},
  doi = {10.48550/arXiv.2506.21539},
  url = {https://arxiv.org/abs/2506.21539}
}

@misc{sun2026vlajepa,
  title = {{VLA-JEPA}: Enhancing Vision-Language-Action Model with Latent World Model},
  author = {Sun, Jingwen and Zhang, Wenyao and Qi, Zekun and Ren, Shaojie and Liu, Zezhi and Zhu, Hanxin and Sun, Guangzhong and Jin, Xin and Chen, Zhibo},
  year = {2026},
  eprint = {2602.10098},
  archivePrefix = {arXiv},
  primaryClass = {cs.RO},
  doi = {10.48550/arXiv.2602.10098},
  url = {https://arxiv.org/abs/2602.10098}
}

@misc{lerobot2026vlajepalibero,
  title = {{VLA-JEPA-LIBERO} Model Repository},
  author = {{LeRobot Team}},
  howpublished = {\url{https://huggingface.co/lerobot/VLA-JEPA-LIBERO}},
  year = {2026},
  note = {Accessed: 2026-07-28}
}

@misc{shukor2025smolvla,
  title = {{SmolVLA}: A Vision-Language-Action Model for Affordable and Efficient Robotics},
  author = {Shukor, Mustafa and Aubakirova, Dana and Capuano, Francesco and Kooijmans, Pepijn and Palma, Steven and Zouitine, Adil and Aractingi, Michel and Pascal, Caroline and Russi, Martino and Marafioti, Andres and Alibert, Simon and Cord, Matthieu and Wolf, Thomas and Cadene, Remi},
  year = {2025},
  eprint = {2506.01844},
  archivePrefix = {arXiv},
  primaryClass = {cs.LG},
  doi = {10.48550/arXiv.2506.01844},
  url = {https://arxiv.org/abs/2506.01844}
}

@misc{lerobot2025smolvlalibero,
  title = {{SmolVLA-LIBERO} Model Card},
  author = {{LeRobot Team}},
  howpublished = {\url{https://huggingface.co/lerobot/smolvla_libero}},
  year = {2025},
  note = {Accessed: 2026-07-28}
}

@misc{beeface2026smolvlaliberospatial,
  title = {{SmolVLA} Fine-Tuned on {LIBERO-Spatial}: Model Card},
  author = {{Beeface}},
  howpublished = {\url{https://huggingface.co/Beeface/smolvla-libero-spatial}},
  year = {2026},
  note = {Accessed: 2026-07-28}
}

@inproceedings{shi2026memoryvla,
  title = {{MemoryVLA}: Perceptual-Cognitive Memory in Vision-Language-Action Models for Robotic Manipulation},
  author = {Shi, Hao and Xie, Bin and Liu, Yingfei and Sun, Lin and Liu, Fengrong and Wang, Tiancai and Zhou, Erjin and Fan, Haoqiang and Zhang, Xiangyu and Huang, Gao},
  booktitle = {International Conference on Learning Representations},
  year = {2026},
  eprint = {2508.19236},
  archivePrefix = {arXiv},
  primaryClass = {cs.RO},
  doi = {10.48550/arXiv.2508.19236},
  url = {https://arxiv.org/abs/2508.19236}
}

@misc{lin2026worldpilot,
  title = {World Pilot: Steering Vision-Language-Action Models with World-Action Priors},
  author = {Lin, Zefu and Cui, Rongxu and Xu, Junjia and Jin, Xiaojuan and Li, Wenling and Fan, Lue and Zhang, Zhaoxiang},
  year = {2026},
  eprint = {2606.12403},
  archivePrefix = {arXiv},
  primaryClass = {cs.RO},
  doi = {10.48550/arXiv.2606.12403},
  url = {https://arxiv.org/abs/2606.12403}
}

@misc{shi2026memoryvlaplus,
  title = {{MemoryVLA++}: Temporal Modeling via Memory and Imagination in Vision-Language-Action Models},
  author = {Shi, Hao and Li, Weiye and Xie, Bin and Wang, Yulin and Zhou, Renping and Wang, Tiancai and Zhang, Xiangyu and Luo, Ping and Huang, Gao},
  year = {2026},
  eprint = {2606.09827},
  archivePrefix = {arXiv},
  primaryClass = {cs.RO},
  doi = {10.48550/arXiv.2606.09827},
  url = {https://arxiv.org/abs/2606.09827}
}

@misc{liu2026goal2skill,
  title = {{Goal2Skill}: Long-Horizon Manipulation with Adaptive Planning and Reflection},
  author = {Liu, Zhen and Ning, Xinyu and Hu, Zhe and Xie, Xinxin and Li, Weize and Tang, Zhipeng and Wang, Chongyu and Yang, Zejun and Wang, Hanlin and Liu, Yitong and Pu, Zhongzhu},
  year = {2026},
  eprint = {2604.13942},
  archivePrefix = {arXiv},
  primaryClass = {cs.RO},
  doi = {10.48550/arXiv.2604.13942},
  url = {https://arxiv.org/abs/2604.13942}
}

@misc{driess2023palme,
  title = {{PaLM-E}: An Embodied Multimodal Language Model},
  author = {Driess, Danny and Xia, Fei and Sajjadi, Mehdi S. M. and Lynch, Corey and Chowdhery, Aakanksha and Ichter, Brian and Wahid, Ayzaan and Tompson, Jonathan and Vuong, Quan and Yu, Tianhe and Huang, Wenlong and Chebotar, Yevgen and Sermanet, Pierre and Duckworth, Daniel and Levine, Sergey and Vanhoucke, Vincent and Hausman, Karol and Toussaint, Marc and Greff, Klaus and Zeng, Andy and Mordatch, Igor and Florence, Pete},
  year = {2023},
  eprint = {2303.03378},
  archivePrefix = {arXiv},
  primaryClass = {cs.LG},
  doi = {10.48550/arXiv.2303.03378},
  url = {https://arxiv.org/abs/2303.03378}
}

@misc{openx2023,
  title = {Open {X}-Embodiment: Robotic Learning Datasets and {RT-X} Models},
  author = {{Open X-Embodiment Collaboration}},
  year = {2023},
  eprint = {2310.08864},
  archivePrefix = {arXiv},
  primaryClass = {cs.RO},
  doi = {10.48550/arXiv.2310.08864},
  url = {https://arxiv.org/abs/2310.08864}
}

@misc{octo2024,
  title = {{Octo}: An Open-Source Generalist Robot Policy},
  author = {{Octo Model Team} and Ghosh, Dibya and Walke, Homer and Pertsch, Karl and Black, Kevin and Mees, Oier and Dasari, Sudeep and Hejna, Joey and Kreiman, Tobias and Xu, Charles and Luo, Jianlan and Tan, You Liang and Chen, Lawrence Yunliang and Sanketi, Pannag and Vuong, Quan and Xiao, Ted and Sadigh, Dorsa and Finn, Chelsea and Levine, Sergey},
  year = {2024},
  eprint = {2405.12213},
  archivePrefix = {arXiv},
  primaryClass = {cs.RO},
  doi = {10.48550/arXiv.2405.12213},
  url = {https://arxiv.org/abs/2405.12213}
}

@misc{liu2024rdt,
  title = {{RDT-1B}: A Diffusion Foundation Model for Bimanual Manipulation},
  author = {Liu, Songming and Wu, Lingxuan and Li, Bangguo and Tan, Hengkai and Chen, Huayu and Wang, Zhengyi and Xu, Ke and Su, Hang and Zhu, Jun},
  year = {2024},
  eprint = {2410.07864},
  archivePrefix = {arXiv},
  primaryClass = {cs.RO},
  doi = {10.48550/arXiv.2410.07864},
  url = {https://arxiv.org/abs/2410.07864}
}

@misc{ahn2022saycan,
  title = {Do As I Can, Not As I Say: Grounding Language in Robotic Affordances},
  author = {Ahn, Michael and others},
  year = {2022},
  eprint = {2204.01691},
  archivePrefix = {arXiv},
  primaryClass = {cs.RO},
  doi = {10.48550/arXiv.2204.01691},
  url = {https://arxiv.org/abs/2204.01691}
}

@misc{huang2022inner,
  title = {Inner Monologue: Embodied Reasoning through Planning with Language Models},
  author = {Huang, Wenlong and Xia, Fei and Xiao, Ted and Chan, Harris and Liang, Jacky and Florence, Pete and Zeng, Andy and Tompson, Jonathan and Mordatch, Igor and Chebotar, Yevgen and Sermanet, Pierre and Brown, Noah and Jackson, Tomas and Luu, Linda and Levine, Sergey and Hausman, Karol and Ichter, Brian},
  year = {2022},
  eprint = {2207.05608},
  archivePrefix = {arXiv},
  primaryClass = {cs.RO},
  doi = {10.48550/arXiv.2207.05608},
  url = {https://arxiv.org/abs/2207.05608}
}

@misc{liang2022code,
  title = {Code as Policies: Language Model Programs for Embodied Control},
  author = {Liang, Jacky and Huang, Wenlong and Xia, Fei and Xu, Peng and Hausman, Karol and Ichter, Brian and Florence, Pete and Zeng, Andy},
  year = {2022},
  eprint = {2209.07753},
  archivePrefix = {arXiv},
  primaryClass = {cs.RO},
  doi = {10.48550/arXiv.2209.07753},
  url = {https://arxiv.org/abs/2209.07753}
}

@misc{huang2023voxposer,
  title = {{VoxPoser}: Composable 3D Value Maps for Robotic Manipulation with Language Models},
  author = {Huang, Wenlong and Wang, Chen and Zhang, Ruohan and Li, Yunzhu and Wu, Jiajun and Fei-Fei, Li},
  year = {2023},
  eprint = {2307.05973},
  archivePrefix = {arXiv},
  primaryClass = {cs.RO},
  doi = {10.48550/arXiv.2307.05973},
  url = {https://arxiv.org/abs/2307.05973}
}

@misc{yang2025instructvla,
  title = {{InstructVLA}: Vision-Language-Action Instruction Tuning from Understanding to Manipulation},
  author = {Yang, Shuai and Li, Hao and Wang, Bin and Chen, Yilun and Tian, Yang and Wang, Tai and Wang, Hanqing and Zhao, Feng and Liao, Yiyi and Pang, Jiangmiao},
  year = {2025},
  eprint = {2507.17520},
  archivePrefix = {arXiv},
  primaryClass = {cs.RO},
  doi = {10.48550/arXiv.2507.17520},
  url = {https://arxiv.org/abs/2507.17520}
}

@misc{liu2026vls,
  title = {{VLS}: Steering Pretrained Robot Policies via Vision-Language Models},
  author = {Liu, Shuo and Singh, Ishneet Sukhvinder and Xu, Yiqing and Duan, Jiafei and Krishna, Ranjay},
  year = {2026},
  eprint = {2602.03973},
  archivePrefix = {arXiv},
  primaryClass = {cs.RO},
  doi = {10.48550/arXiv.2602.03973},
  url = {https://arxiv.org/abs/2602.03973}
}

@misc{srikanth2026qdig,
  title = {Red-Teaming Vision-Language-Action Models via Quality Diversity Prompt Generation for Robust Robot Policies},
  author = {Srikanth, Siddharth and Liang, Freddie and Hsu, Ya-Chuan and Bhatt, Varun and Zhao, Shihan and Chen, Henry and Tjanaka, Bryon and Hwang, Minjune and Saran, Akanksha and Seita, Daniel and Tabrez, Aaquib and Nikolaidis, Stefanos},
  year = {2026},
  eprint = {2603.12510},
  archivePrefix = {arXiv},
  primaryClass = {cs.RO},
  doi = {10.48550/arXiv.2603.12510},
  url = {https://arxiv.org/abs/2603.12510}
}

@misc{xie2026strongvla,
  title = {{STRONG-VLA}: Decoupled Robustness Learning for Vision-Language-Action Models under Multimodal Perturbations},
  author = {Xie, Yuhan and Yan, Yuping and Zhao, Yunqi and Wang, Handing and Jin, Yaochu},
  year = {2026},
  eprint = {2604.10055},
  archivePrefix = {arXiv},
  primaryClass = {cs.RO},
  doi = {10.48550/arXiv.2604.10055},
  url = {https://arxiv.org/abs/2604.10055}
}

@misc{wang2026wamsurvey,
  title = {World Action Models: The Next Frontier in Embodied {AI}},
  author = {Wang, Siyin and Shi, Junhao and Fu, Zhaoyang and He, Xinzhe and Liu, Feihong and Yang, Chenchen and Zhou, Yikang and Fei, Zhaoye and Gong, Jingjing and Fu, Jinlan and Shou, Mike Zheng and Huang, Xuanjing and Qiu, Xipeng and Jiang, Yu-Gang},
  year = {2026},
  eprint = {2605.12090},
  archivePrefix = {arXiv},
  primaryClass = {cs.RO},
  doi = {10.48550/arXiv.2605.12090},
  url = {https://arxiv.org/abs/2605.12090}
}

@misc{garrido2026latentaction,
  title = {Learning Latent Action World Models in the Wild},
  author = {Garrido, Quentin and Nagarajan, Tushar and Terver, Basile and Ballas, Nicolas and LeCun, Yann and Rabbat, Michael},
  year = {2026},
  eprint = {2601.05230},
  archivePrefix = {arXiv},
  primaryClass = {cs.CV},
  doi = {10.48550/arXiv.2601.05230},
  url = {https://arxiv.org/abs/2601.05230}
}

@misc{brohan2022rt1,
  title = {{RT-1}: Robotics Transformer for Real-World Control at Scale},
  author = {Brohan, Anthony and others},
  year = {2022},
  eprint = {2212.06817},
  archivePrefix = {arXiv},
  primaryClass = {cs.RO},
  doi = {10.48550/arXiv.2212.06817},
  url = {https://arxiv.org/abs/2212.06817}
}

@misc{chen2025fastinslow,
  title = {Fast-in-Slow: A Dual-System Foundation Model Unifying Fast Manipulation within Slow Reasoning},
  author = {Chen, Hao and Liu, Jiaming and Gu, Chenyang and Liu, Zhuoyang and Zhang, Renrui and Li, Xiaoqi and He, Xiao and Guo, Yandong and Fu, Chi-Wing and Zhang, Shanghang and Heng, Pheng-Ann},
  year = {2025},
  eprint = {2506.01953},
  archivePrefix = {arXiv},
  primaryClass = {cs.RO},
  doi = {10.48550/arXiv.2506.01953},
  url = {https://arxiv.org/abs/2506.01953}
}

@misc{zou2025duocore,
  title = {Asynchronous Fast-Slow Vision-Language-Action Policies for Whole-Body Robotic Manipulation},
  author = {Zou, Teqiang and Zeng, Hongliang and Nong, Yuxuan and Li, Yifan and Liu, Kehui and Yang, Haotian and Ling, Xinyang and Li, Xin and Ma, Lianyang},
  year = {2025},
  eprint = {2512.20188},
  archivePrefix = {arXiv},
  primaryClass = {cs.RO},
  doi = {10.48550/arXiv.2512.20188},
  url = {https://arxiv.org/abs/2512.20188}
}

@inproceedings{sclar2024promptformat,
  title = {Quantifying Language Models' Sensitivity to Spurious Features in Prompt Design or: How I Learned to Start Worrying about Prompt Formatting},
  author = {Sclar, Melanie and Choi, Yejin and Tsvetkov, Yulia and Suhr, Alane},
  booktitle = {International Conference on Learning Representations},
  year = {2024},
  url = {https://proceedings.iclr.cc/paper_files/paper/2024/hash/6c0e99d736da621403018ca7b32b1a4d-Abstract-Conference.html}
}

@inproceedings{webson2022prompts,
  title = {Do Prompt-Based Models Really Understand the Meaning of Their Prompts?},
  author = {Webson, Albert and Pavlick, Ellie},
  booktitle = {Proceedings of the 2022 Conference of the North American Chapter of the Association for Computational Linguistics: Human Language Technologies},
  pages = {2300--2344},
  publisher = {Association for Computational Linguistics},
  year = {2022},
  doi = {10.18653/v1/2022.naacl-main.167},
  url = {https://aclanthology.org/2022.naacl-main.167/}
}

@article{zou2023universal,
  title = {Universal and Transferable Adversarial Attacks on Aligned Language Models},
  author = {Zou, Andy and Wang, Zifan and Carlini, Nicholas and Nasr, Milad and Kolter, J. Zico and Fredrikson, Matt},
  journal = {arXiv preprint arXiv:2307.15043},
  year = {2023},
  url = {https://arxiv.org/abs/2307.15043}
}

@inproceedings{alshiekh2018shielding,
  title = {Safe Reinforcement Learning via Shielding},
  author = {Alshiekh, Mohammed and Bloem, Roderick and Ehlers, R{\"u}diger and K{\"o}nighofer, Bettina and Niekum, Scott and Topcu, Ufuk},
  booktitle = {Proceedings of the AAAI Conference on Artificial Intelligence},
  volume = {32},
  year = {2018},
  doi = {10.1609/aaai.v32i1.11797},
  url = {https://ojs.aaai.org/index.php/AAAI/article/view/11797}
}
}

\end{document}